%% file: main.tex
\documentclass{article}

\usepackage{microtype}
\usepackage{graphicx}
\usepackage{subcaption}
\usepackage{booktabs} 
\usepackage{url}
\usepackage{tikz}
\usepackage{pgfplots}
\pgfplotsset{compat=1.18} 

\usepackage{hyperref}

\usepackage[preprint]{icml2026}

\usepackage{amsmath}
\usepackage{amssymb}
\usepackage{mathtools}
\usepackage{amsthm}
\usepackage{booktabs}
\usepackage{multirow}
\usepackage[table]{xcolor}
\usepackage[most]{tcolorbox}
\usepackage{xcolor}
\usepackage[capitalize,noabbrev]{cleveref}
\usepackage{float}
\theoremstyle{plain}

\theoremstyle{definition}

\theoremstyle{remark}

\usepackage[textsize=tiny]{todonotes}

\newcommand{\tocite}[1]{\textcolor{green}{(cite)}}

\definecolor{PromptGray}{RGB}{120,120,120}
\definecolor{PromptBlue}{RGB}{48,98,166}
\definecolor{PromptBrown}{RGB}{139,94,60}
\definecolor{PromptTeal}{RGB}{46,125,122}
\tcbset{
  promptbox/.style={
    enhanced,
    colback=white,
    colframe=black!25,
    boxrule=2.0pt,
    arc=2pt,
    outer arc=2pt,
    left=6pt,right=6pt,top=6pt,bottom=6pt,
    fonttitle=\bfseries\small,
    coltitle=white,
    attach boxed title to top left={xshift=6pt,yshift=-1.5mm},
    boxed title style={
      boxrule=0pt,
      arc=2pt,
      outer arc=2pt,
      left=5pt,right=5pt,top=2pt,bottom=2pt
    },
  }
}
\usepackage{enumitem}
\setlist[description]{leftmargin=1.2em,labelsep=0.5em,itemsep=0pt,topsep=2pt,parsep=0pt}

\newtcolorbox{PromptGrayBox}[1]{promptbox,
  title=#1,
  colframe=PromptGray,
  boxed title style={colback=PromptGray}
}

\newtcolorbox{PromptBlueBox}[1]{promptbox,
  title=#1,
  colframe=PromptBlue,
  boxed title style={colback=PromptBlue}
}

\newtcolorbox{PromptBrownBox}[1]{promptbox,
  title=#1,
  colframe=PromptBrown,
  boxed title style={colback=PromptBrown}
}

\newtcolorbox{PromptTealBox}[1]{promptbox,
  title=#1,
  colframe=PromptTeal,
  boxed title style={colback=PromptTeal, colframe=PromptTeal}
}

\icmltitlerunning{FlexRec: Adapting LLM-based Recommenders for Flexible Needs via Reinforcement Learning}

\begin{document}
\twocolumn[
  \icmltitle{FlexRec: Adapting LLM-based Recommenders for \\Flexible Needs via Reinforcement Learning}



  \icmlsetsymbol{adv}{*}

  \begin{icmlauthorlist}
    \icmlauthor{Yijun Pan}{yale}
    \icmlauthor{Weikang Qiu}{yale}
    \icmlauthor{Qiyao Ma}{ucd}
    \icmlauthor{Mingxuan Ju}{snap,adv}
    \icmlauthor{Tong Zhao}{snap,adv}
    \icmlauthor{Neil Shah}{snap,adv}
    \icmlauthor{Rex Ying}{yale}
  \end{icmlauthorlist}
    \icmlcorrespondingauthor{Rex Ying}{rex.ying@yale.edu}
  \icmlaffiliation{yale}{Yale University}
  \icmlaffiliation{snap}{Snap Inc}
\icmlaffiliation{ucd}{University of California, Davis}

  \icmlkeywords{Machine Learning, ICML}

  \vskip 0.3in
]



\printAffiliationsAndNotice{*Authors affiliated with Snap Inc. served in advisory roles only for this work.}  

\begin{abstract}
Modern recommender systems must adapt to dynamic, need-specific objectives for diverse recommendation scenarios, yet most traditional recommenders are optimized for a single static target and struggle to reconfigure behavior on demand. Recent advances in reinforcement-learning-based post-training have unlocked strong instruction-following and reasoning capabilities in LLMs, suggesting a principled route for aligning them to complex recommendation goals. Motivated by this, we study closed-set autoregressive ranking, where an LLM generates a permutation over a fixed candidate set conditioned on user context and an explicit need instruction. However, applying RL to this setting faces two key obstacles: (i) sequence-level rewards yield coarse credit assignment that fails to provide fine-grained training signals, and (ii) interaction feedback is sparse and noisy, which together lead to inefficient and unstable updates. We propose FlexRec, a post-training RL framework that addresses both issues with (1) a causally grounded item-level reward based on counterfactual swaps within the remaining candidate pool, and (2) critic-guided, uncertainty-aware scaling that explicitly models reward uncertainty and down-weights low-confidence rewards to stabilize learning under sparse supervision. Across diverse recommendation scenarios and objectives, FlexRec achieves substantial gains: it improves NDCG@5 by up to \textbf{59\%} and Recall@5 by up to \textbf{109.4\%} in need-specific ranking, and further achieves up to \textbf{24.1\%} Recall@5 improvement under generalization settings, outperforming strong traditional recommenders and LLM-based baselines.
\end{abstract}

\input{01_introduction}

\input{02_related_work}
\input{03_method_item_level}

\input{03_method_critic}

\input{04_experiments}

\input{05_conclusion}

\bibliography{reference}
\bibliographystyle{icml2026}

\newpage
\input{10_appendix}
\end{document}

%% file: 01_introduction.tex
    

\section{Introduction}

Existing recommender systems are typically optimized for a single objective (e.g., click-through rate in advertising or purchases in e-commerce platforms \cite{he2017neuralfactorizationmachinessparse, Guo2017DeepFMAF}). While exhibiting high performance, this paradigm fails to reflect the complexity of real-world recommendation scenarios. In practice, user intents are inherently dynamic \cite{mao2024analysis,systems13111034}. Meanwhile, merchants' objectives can also shift over time, for example, between brand awareness and driving sales. The inability to flexibly adapt recommendation strategies to such evolving user and business needs remains a fundamental limitation of current recommender systems.

LLMs offer a promising direction for addressing this challenge. With strong instruction-following and generalization abilities \cite{wei2022chain,wang2025multimodalchainofthoughtreasoningcomprehensive,brown2020language,gao2021making}, they provide a natural interface for modeling complex preferences and adapting recommendation behavior through prompts or task specifications. This motivates the integration of LLMs into recommender systems \cite{deng2025onerec,lin2025can,zhang2023user}, potentially enabling dynamic control over recommendation objectives without retraining task-specific models. However, because LLMs are predominantly trained on large-scale general-purpose text corpora, they remain weakly aligned with recommendation tasks and cannot be directly deployed as recommenders without further adaptation.

Reinforcement learning from verifiable rewards (RLVR) \cite{lambert2024tulu} has recently emerged as an effective paradigm for aligning LLMs with downstream objectives. Recent work has begun applying RLVR to LLM-based recommendation. Rec-R1 \cite{lin2025recr1bridginggenerativelarge} proposes a general framework that optimizes an LLM with sequence-level feedback from downstream recommendation models using GRPO \cite{shao2024deepseekmathpushinglimitsmathematical}, while ConvRec-R1 \cite{zhu2025rankgrpotrainingllmbasedconversational} further explores rank-level GRPO for conversational recommendation.

However, adopting GRPO for dynamic recommendation introduces several fundamental challenges. First, interaction annotations are inherently sparse. In realistic scenarios, only a small subset of items receives explicit feedback from the user, while the majority remain unlabeled. Existing work therefore often relies on learned interaction predictors to complete missing supervision \cite{yang2025enhancelargelanguagemodels,wang2025searchbasedinteractionconversationrecommendation}, but these estimates inevitably contain error and can vary substantially in reliability. This is especially problematic in RL, where high-variance reward estimates can strongly distort advantage estimation and destabilize policy updates \cite{zhu2025vrporethinkingvaluemodeling}. 

Second, existing RL methods for LLM-based recommendation systems typically operate with sequence-level rewards \cite{shao2024deepseekmathpushinglimitsmathematical,deng2025onerec,wu2025rlpf}, assigning a single scalar value to the entire ranked list. However, recommendation is a structured decision process in which the model constructs a ranking through a sequence of atomic item placements. A single list-level reward therefore fails to capture which atomic ranking decisions are beneficial or harmful, obscuring fine-grained credit assignment despite the richer supervision inherently available in ranked outputs. Although some prior work \cite{zhu2025rankgrpotrainingllmbasedconversational} explores item-level supervision, the reward design makes item-level rewards not directly comparable across items, leading to unstable updates. 



To address these challenges, we propose FlexRec, a principled post-training framework for aligning LLM-based recommender systems with dynamic recommendation strategies. Our key observation is that RL for recommendation must simultaneously solve two problems: \textbf{fine-grained credit assignment} over a ranked list and \textbf{robust optimization} under sparse, noisy reward completion. 
Accordingly, our approach makes two key contributions:
\begin{itemize}
  \item \textbf{Uncertainty-guided GRPO.} To address the challenge of sparse and noisy reward completion, we train a critic that jointly predicts reward values and their uncertainty, and integrate this confidence signal directly into GRPO updates to down-weight unreliable reward estimates.
  \item \textbf{Swap-based item-level reward.} To address the challenge of coarse sequence-level supervision, we introduce a counterfactual swap operation within the remaining candidate pool to evaluate the marginal contribution of each item placement, producing dense, position-aware supervision for autoregressive ranking.
\end{itemize}

Together, these components enable stable and effective alignment of LLMs for dynamic, need-aware recommendation. Our experiments across diverse recommendation scenarios and user needs: including maximizing interest, encouraging exploration, and promoting trending items, demonstrate that FlexRec effectively aligns LLMs as strong recommenders, outperforming both traditional and LLM-based baselines. We further show that the trained LLM recommender generalizes efficiently to unseen needs. Finally, a single LLM trained jointly on all needs remains competitive across all scenarios, dynamically adapting its reasoning and ranking strategy, making it practical as a universal recommender.



%% file: 02_related_work.tex
\section{Related Work}

\paragraph{LLMs for Multi-Need Recommendation}
Recommender systems have shifted from traditional collaborative filtering and ID-based sequential encoders (e.g., BERT4Rec \cite{sun2019bert4rec} and S3-Rec \cite{zhou2020s3}) toward generative paradigms built on Large Language Models (LLMs). Current approaches typically rely on instruction-tuning \cite{deng2025onerec, lin2025recr1bridginggenerativelarge, rajput2023recommender, wang2024eager} or in-context learning \cite{lyu2024llm} to leverage LLMs' semantic reasoning and world knowledge for modeling user behavior. However, most LLM-based recommenders remain \emph{static evaluators} of historical data \cite{wu2018personalizing, dhelim2022survey, ma2024breaking, zhang2023user, yang2023palr}: they optimize for a single implicit objective and can struggle when a user’s immediate, real-time requirements deviate from their long-term profile. This motivates recommendation models that can condition on explicit need specifications and adapt their behavior accordingly. In this work, we take a step toward multi-need LLM recommendation by contributing a suite of need-specific instruction-tuning datasets tailored for LLM post-training, and by proposing FlexRec, which adapts to flexible user needs at inference.


\paragraph{Reinforcement Learning from Verifiable Rewards}
Reinforcement Learning from Verifiable Rewards (RLVR) has significantly enhanced LLM reasoning in deterministic domains such as mathematics \cite{shao2024deepseekmathpushinglimitsmathematical}, coding \cite{chen2021evaluating}, and open-domain tasks \cite{ma2025faithfulness}. 
Recent RLVR-based post training methods for recommendation like Rec-R1 use sequence-level metrics such as NDCG \cite{lin2025recr1bridginggenerativelarge} to guide model training. While effective, this design assigns a single advantage to all tokens in a rollout, yielding coarse credit assignment that cannot distinguish the contributions of individual items. ConvRec-R1 \citet{zhu2025rankgrpotrainingllmbasedconversational} use a rank-level reward; however, in autoregressive ranking the reward at a given rank depends on the prefix and remaining candidate pool, so it is not comparable across rollouts and can bias credit assignment. In contrast, FlexRec introduces a swap-based item-level reward that respects the fundamental properties of autoregressive ranking to enable fine-grained updates, leading to faster convergence and better performance.



\paragraph{Sparse User Interactions}
Data sparsity poses a major challenge in recommendation: most user--item pairs are unobserved, making RLVR difficult due to limited reward supervision. A common remedy is to train a lightweight neural critic to impute rewards for unobserved pairs \cite{lobel2020amortizedrankingcriticaltrainingcollaborative,10.1145/3038912.3052569,Guo2017DeepFMAF,he2017neuralfactorizationmachinessparse}, but proxy errors are particularly harmful for LLM post-training since erroneous advantage (magnitude and sign) directly drives policy updates. To mitigate this, FlexRec augments the critic to predict both value and uncertainty, and incorporates uncertainty into the update to down-weight unreliable estimates and stabilize training.



%% file: 03_method_item_level.tex
\section{Method}
In this section, we present FlexRec, an RL post-training framework that trains LLM recommenders via item-level GRPO and uncertainty-aware updates under sparse feedback.

\begin{figure*}
    \centering
    \includegraphics[width=0.9\linewidth]{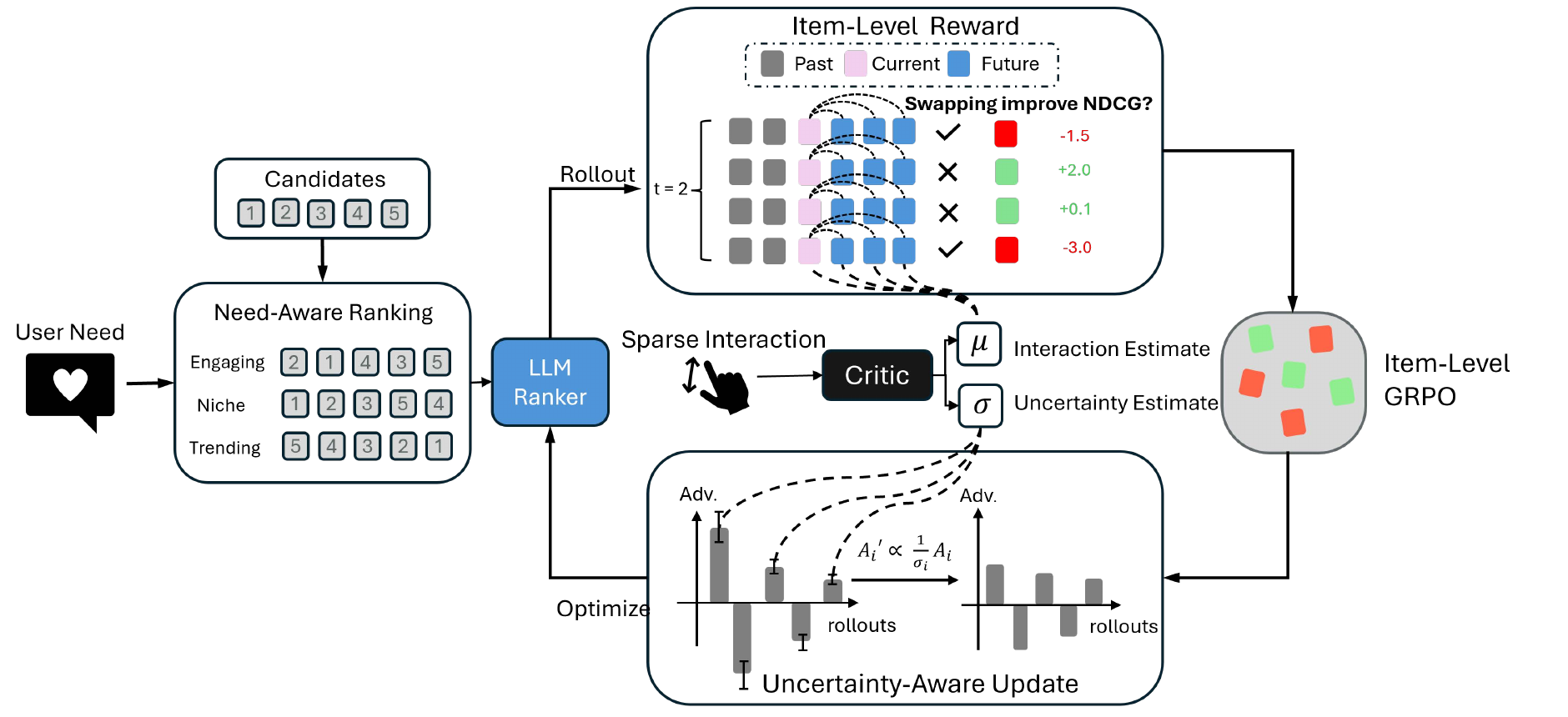}
    \caption{Overall framework of FlexRec. Given a candidate set and an explicit user need, an LLM recommender generates ranked rollouts. An item-level reward is computed by evaluating the marginal contribution of individual item placements via counterfactual swaps (top right). A critic predicts both the expected interaction reward and its uncertainty for unobserved interactions. These estimates are used to form uncertainty-aware advantages, down-weighting unreliable signals during optimization (bottom). 
    }
    \label{fig:framework}
\end{figure*}

\subsection{Problem Formulation: Need-specific Autoregressive Ranking}
We study need-specific ranking, where the model outputs an ordered list optimized for a given need, and each need induces a relevance function over the candidate set. We represent both context and candidate items in natural language and produce rankings via text generation, enabling LLMs to reason over heterogeneous signals (e.g., user history, item metadata, and need instructions) and to transfer more easily to new needs without relying on task-specific retraining. 

Let $x \triangleq (U, C, M)$ denote the context, where $U$ is user information (e.g., history), $C$ is the candidate set and $M$ is metadata/task instruction.
Let $n \in \mathcal{N}$ denote the need (e.g., engagement, niche discovery, trend promotion).


Given a candidate set $C$, a language model induces a need-conditioned ranking policy by generating an ordered sequence over $C$ (with optional reasoning)
\[
y = (a_1,\dots,a_K), 
\qquad a_k \in C \setminus \{a_1,\dots,a_{k-1}\},
\]
where at each step $k$ it selects one item from the remaining candidate pool. This defines an autoregressive decision process $\pi_\theta(a_k \mid a_{<k}, x, n)$
which maps the context, the target need, and the previously selected items $a_{<k}$ to a distribution over the remaining candidates.

The quality of a generated ranking $y$ is evaluated by a need-specific objective $R_n(y;x)$, which is decomposable over items: 
\[
    R_n(y;x) \;=\; \sum_{k=1}^{K} r_n(a_k;x), \forall n
\]

where $r_n(a_k\,;\ x)$ is the position-dependent gain of placing item $a_k$ at rank $k$ under context $x$ and need $n$ (e.g. click label).
Following previous work \cite{lin2025recr1bridginggenerativelarge,zhu2025rankgrpotrainingllmbasedconversational}, we instantiate $R_n(y\,;\,x)$ as NDCG @ K.


\subsection{Swap-based Item-level Reward}




The primary limitations of prior RL-based LLM recommenders are reward designs that are either overly coarse or biased: sequence-level GRPO like Rec-R1 \cite{lin2025recr1bridginggenerativelarge} assigns a single advantage to all tokens, while per-position gains used in ConvRec-R1 \cite{zhu2025rankgrpotrainingllmbasedconversational} depend on the prefix and remaining candidate pool, making item-level signals incomparable across rollouts and prone to inaccurate credit assignment.



\paragraph{Properties of Autoregressive Reranking.}
To address these limitations, we formalize three fundamental properties of autoregressive ranking with language models. These insights characterize the structure of the ranking problem and guide the design of correct item-level reward signals.


\noindent\textbf{Insight 1:  Causality.}
The contribution of $a_k$ is defined only relative to the remaining pool
$\mathcal{C}_k = C \setminus \{a_1,\dots,a_{k-1}\}$, because earlier-ranked items are already fixed and cannot be affected by the choice of $a_k$.
Therefore, any valid credit signal for $a_k$ should depend on how it compares to items in
$\mathcal{C}_k$, rather than on its absolute contribution alone.

\noindent\textbf{Insight 2: Swaps as Atomic Actions.}
Let $C=\{c_1,\dots,c_K\}$ and let a ranking be a permutation $y=(a_1,\dots,a_K)\in S_K$.
For any pair of positions $1\le k<j\le K$, define the \emph{swap operator} $\tau_{k,j}$ that exchanges the items at ranks $k$ and $j$, and denote the swapped ranking by
\[
y^{(k\leftrightarrow j)} \;\triangleq\; \tau_{k,j}(y).
\]
The set $\{\tau_{k,j}\}_{k<j}$ generates $S_K$, i.e., any permutation can be obtained by composing pairwise swaps. We therefore view a single swap as an atomic local edit of a ranking, and attribute local changes in listwise quality to swapping.

\noindent\textbf{Insight 3:  Marginal Contribution via Counterfactual Relocation.}
Item-level credit should be defined through counterfactual rankings in which the position of an item is changed while the rest of the list is held fixed. Using the swap operator from insight 2, the natural local counterfactual for the item at rank $k$ is to exchange it with a lower-ranked item $j>k$, yielding $y^{(k\leftrightarrow j)}$. We therefore quantify the contribution of $a_k$ by the change in objective:
\[
\Delta_{k,j}(y;x) \;\triangleq\; R_n\!\left(y^{(k\leftrightarrow j)};x\right) - R_n(y;x),
\]
and define the item-level reward as the expected improvement under an atomic swap over the remaining positions.


\paragraph{Reward Design}
Motivated by the previous insights, we define item-level credit through counterfactual swaps.  
Given a predicted ranking $y=(a_1,\dots,a_K)\in S_K$ and a need-specific listwise objective $R_n(y;x)$, let
$y^{(k\leftrightarrow j)}$ denote the ranking obtained by swapping positions $k$ and $j$ ($j>k$).
For the item placed at rank $k$, we define its improvement as
\begin{equation*}\label{eq:delta_t}
\Delta_k(y;x)\triangleq
\mathbb{E}_{j\sim\mathrm{Unif}([k{+}1{:}K])}
\!\left[\,R_n\!\big(y^{(k\leftrightarrow j)};x\big)-R_n(y;x)\right].
\end{equation*}

which compares $a_k$ only against items in the remaining candidate pool at step $k$. We estimate $\Delta_k$ by averaging over all such swaps, yielding the item-level reward
\begin{equation}
r_k^{\mathrm{CS}}
\;=\;
-\frac{1}{K-k}
\sum_{j=k+1}^{K}
\Big( R_n\!\left(y^{(k\leftrightarrow j)};x\right) - R_n(y;x) \Big).
\label{eq:causalswap}
\end{equation}
By construction, $r_k^{\mathrm{CS}}$ is an unbiased estimator of $\Delta_k$.

\paragraph{Time Complexity}

Our calculation of item-level reward introduced $\mathcal{O}(K^2)
$computation overhead for K ranked items, which is minimal given the limited size of candidates in the ranking stage.

\paragraph{Item-Level GRPO}

Given item-level rewards, we compute advantages by normalizing across rollouts within each prompt group.
Let $g$ index a prompt group and let $y^{(i)}=(a^{(i)}_1,\dots,a^{(i)}_K)$ denote the $i$-th rollout in group $g$.
Let $r^{\mathrm{CS}}_{g,i,k}$ be the item-level reward (Eq.~\eqref{eq:causalswap}) computed on $y^{(i)}$.
We pool all item rewards in group $g$ as
\[
\mathcal{R}_g \triangleq \{\, r^{\mathrm{CS}}_{g,i,k} \mid i=1,\dots,G_g,\; k=1,\dots,K \,\},
\]
and compute group-specific statistics
\[
\mu_g \triangleq \mathbb{E}_{r\sim \mathcal{R}_g}[r],
\qquad
\sigma_g \triangleq \sqrt{\mathrm{Var}_{r\sim \mathcal{R}_g}[r]}.
\]
The item-level advantage is then
\[
A^{\mathrm{item}}_{g,i,k}
\;=\;
\frac{r^{\mathrm{CS}}_{g,i,k}-\mu_g}{\sigma_g+\varepsilon},
\]
which we assign to all tokens corresponding to item $a^{(i)}_k$ in the generated sequence. Unlike standard GRPO, which normalizes sequence-level rewards across rollouts, our normalization is performed over all item rewards across all rollouts. This is well-defined because our design yields comparable, unbiased item-level rewards across varying prefixes and remaining candidate pools, enabling meaningful normalization across contexts.


\paragraph{Sequence-Level Advantages for Non-Item Tokens}

Not all tokens in a generated sequence correspond to item predictions. Reasoning, formatting, and control tokens affect the overall list quality rather than any single position. For these tokens, we use standard sequence-level GRPO within each prompt group.

Let $g$ index a prompt group and let $R_{g,i}$ denote the sequence-level reward of rollout $i$ in group $g$. We compute group statistics
\[
\mu^{\mathrm{seq}}_g \triangleq \mathbb{E}_{i}[R_{g,i}],
\qquad
\sigma^{\mathrm{seq}}_g \triangleq \sqrt{\mathrm{Var}_{i}[R_{g,i}]},
\]
and define the sequence-level advantage
\[
A^{\mathrm{seq}}_{g,i} \;=\; \frac{R_{g,i}-\mu^{\mathrm{seq}}_g}{\sigma^{\mathrm{seq}}_g+\varepsilon}.
\]
We assign $A^{\mathrm{seq}}_{g,i}$ to all non-item tokens in rollout $i$.


Overall, FlexRec provides fine-grained credit for items via while retaining sequence-level guidance for reasoning/formatting tokens via. This hybrid assignment improves learning efficiency (see Section \ref{ablation:reward}) over uniform sequence-level rewards, and avoids the rollout-dependent bias of prior rank-level designs.

%% file: 03_method_critic.tex
\subsection{Uncertainty-Aware GRPO}

As discussed above, sparsity in user–item interactions makes reward imputation in RLVR inherently difficult, so prior work often relies on a critic, typically a lightweight neural model, to predict missing rewards. However, with limited supervision, critic estimates are noisy and often miscalibrated, leading to unstable updates when used for training. To address this challenge, we propose an uncertainty-aware GRPO that down-weights unreliable rewards, preventing spurious estimates from misleading policy updates.

Concretely, we train a neural critic to jointly predict user–item interaction rewards and their \textbf{uncertainty} (see Appendix~\ref{appendix:critic} for architecture and training details). Given context $x$ and a candidate item $a_k$, the critic outputs a mean reward estimate $\hat r_n(a_k; x)$ together with a predictive variance $\mathrm{Var}[r_n(a_k; x)]$, where larger variance indicates less reliable reward estimation. 

Empirically, using critic-predicted rewards yields consistent gains over simple collaborative signals in our ablations (Sec.~\ref{ablat:uncertainty}), underscoring the importance of a learned critic under sparse user--item interactions.

\paragraph{Uncertainty of Sequence Reward}
Consider a predicted ranking $y=(a_1,\dots,a_   K)$. The sequence-level objective could be written as
\begin{equation}
R_n(y;x) \;=\; \sum_{k=1}^{K} \, r_n(a_k ; x),
\end{equation}
where $r_n(a_k ; x)$ denotes the item-level contribution of selecting item $a_k$ under context $x$ and need $n$.
When item-level rewards are estimated rather than observed, the sequence-level objective inherits uncertainty from its components. 
Let $\mathrm{Var}[r_n(a_k;x)]$ denote the uncertainty of the item-level contribution at position $k$ predicted by the critic. 
Under additive aggregation and ignoring cross-item covariance, the uncertainty of the sequence-level reward is
\begin{equation}
\mathrm{Var}[R_n(y;x)] \;=\; \sum_{k=1}^{K} \,\mathrm{Var}[r(a_k;x)].
\end{equation}

\paragraph{Uncertainty-aware Update.}

When rewards are estimated by a noisy critic, the comparison between them remain informative on average, but the confidence of each estimate varies substantially across rollouts. We therefore treat reward uncertainty as a reliability signal that scales the effective advantage, reducing the impact of high-variance estimates while preserving their directional guidance.

Let $A_i$ denote the standard sequence-level advantage for rollout $i$, and let $v_i$ be the estimated variance of its sequence-level reward. To down-weight high-variance updates, we reweigh the advantage by inverse variance:
\[
c_i \triangleq \frac{1}{v_i+\varepsilon},
\qquad
c_i \;\leftarrow\; \min\!\left(\frac{c_i}{\frac{1}{|g(i)|}\sum_{j\in g(i)} c_j},\,1\right).
\]
This reweighting mitigates erroneous updates from a noisy critic: spurious, overconfident reward estimates can induce large advantages with the wrong sign. Finally, we apply the (normalized and clipped) weight to the advantage:
\[
\widetilde A_i \;=\; c_i\,A_i.
\]


As shown in figure \ref{fig:framework}, FlexRec introduces a post-training framework that reliably trains LLMs as recommenders under sparse user--item feedback. It makes two key contributions: (1) a swap-based, item-level reward that provides fine-grained credit assignment for each item, improving learning efficiency; and (2) an uncertainty-aware update that models reward variance to downweight unreliable critic signals, stabilizing RL optimization in sparse settings.


%% file: 04_experiments.tex
\section{Experiments}
In this section, we describe the datasets used in our work and how we construct a diverse suite of needs. We then show that FlexRec is both an efficient learner when trained for a single need and a strong generalist that transfers effectively across needs, enabling a universal LLM ranker (see Appendix \ref{app:rl_training_details} for training details).
\subsection{Datasets}

In this work we focus on the following datasets: KuaiRec, ML-1M and ESCI.

\begin{itemize}[noitemsep,topsep=0pt,leftmargin=*]
  \item \textbf{KuaiRec}~\cite{gao2022kuairec}. A fully observed short-video recommendation dataset with dense user--item interactions and rich user/video metadata; to better reflect sparse implicit feedback, we uniformly subsample 10\% interactions per user.
  \item \textbf{MovieLens} ~\cite{10.1145/3511808.3557643}. A classic rating-based benchmark capturing long-term user preferences over movies.
  \item \textbf{ESCI} ~\citep{reddy2022shopping}. A product-search benchmark where the model ranks a fixed set of items relevant to a textual query.
\end{itemize}
Together, these benchmarks span a wide range of recommendation scenarios, from short-form content consumption and long-horizon preference modeling to query-driven ranking.

\noindent\textbf{Implementation details.} For details of dataset construction refer to Appendix~\ref{appendix:data_details}.



\subsection{Need-Specific Objective Design} 

To address the lack of public benchmarks that explicitly annotate user needs in recommendation, we construct a suite of need-specific objectives on top of each dataset. These needs span realistic scenarios, from maximizing user interest, to surfacing new topics a user is likely to appreciate, to promoting items that are trending on the platform. For details of label construction see Appendix \ref{appendix:need_construction}

\paragraph{Maximizing Interest}
Ranks items by user preference signals, so items the user is most likely to enjoy and engage with receive the highest scores.

\paragraph{Explore New Topics}   
Prioritizes niche or underexposed items that match the user’s interests but come from topics/genres the user has not recently interacted with, so items with strong interest alignment and low historical exposure are scored highest.

\paragraph{Trend Promotion}
Balances personalization with temporal popularity, so items that are both relevant to the user and highly popular in the recent time window (e.g., the past day) receive the highest scores.


\subsection{Evaluation}

\paragraph{Metrics} We evaluate the generated rankings using NDCG at $K \in \{5,10,30\}$ (denoted N@5, N@10, and N@30), along with Recall@5 and MRR@5. These metrics jointly capture overall listwise ranking quality and the quality of the top-ranked items, which is a primary concern in practical recommendation systems. For more details regarding evaluation see Appendix \ref{appendix:evaluation_details}.

\subsection{Baselines}

In this work, we consider a diverse set of baselines, as the closed-set reranking task can be approached using a variety of modeling paradigms. Specifically, we evaluate the following methods:

\paragraph{Traditional Rerankers}
\begin{itemize}[noitemsep,topsep=0pt,leftmargin=*]
  \item \textbf{BERT4Rec}~\cite{sun2019bert4rec}. A bidirectional Transformer sequential recommender trained with an MLM (Masked Language Modeling) objective. We use the standard RecBole implementation.\footnote{\url{https://github.com/RUCAIBox/RecBole/}}
  \item \textbf{STAR}~\cite{lee2025starsimpletrainingfreeapproach}. Uses LLM-based embeddings for sequential recommendation and achieves strong performance without task-specific training. We follow the official implementation in the original paper.
\end{itemize}

\paragraph{Post-trained LLMs}
\begin{itemize}[noitemsep,topsep=0pt,leftmargin=*]
  \item \textbf{TALLRec} ~\cite{Bao_2023}. Applies LoRA fine-tuning using supervised recommendation data; we use the optimal ranked list as the target.
  \item \textbf{Rec-R1}~\cite{lin2025recr1bridginggenerativelarge}. Uses vanilla GRPO with sequence-level ranking rewards (e.g., NDCG). We adopt the official implementation \footnote{\url{https://github.com/linjc16/Rec-R1/tree/main/src}} and apply 25 warm-up SFT steps.
  \item \textbf{ConvRec-R1}~\cite{zhu2025rankgrpotrainingllmbasedconversational}. Extends Rec-R1 with rank-based GRPO for closed-set recommendation, grouping identical rank predictions across rollouts rather than across samples.
\end{itemize}

\subsection{Results}

\paragraph{FlexRec is an Effective Single-Need Recommender}

\begin{table*}[!h]
\small
\centering
\caption{Performance on \textbf{Maximizing Interest} across KuaiRec and MovieLens-1M. FlexRec achieves superior single-need ranking performance compared with baselines.}

\resizebox{1.0 \linewidth}{!}{%
\begin{tabular}{l|ccccc|ccccc}
\toprule
 & \multicolumn{5}{c|}{\textbf{KuaiRec}} & \multicolumn{5}{c}{\textbf{MovieLens-1M}} \\
\textbf{Maximizing Interest}
& \textbf{N@5} & \textbf{N@10} & \textbf{N@30} & \textbf{R@5} & \textbf{MRR@5}
& \textbf{N@5} & \textbf{N@10} & \textbf{N@30} & \textbf{P@5} & \textbf{MRR@5} \\
\midrule

\multicolumn{11}{c}{\textbf{Traditional Rerankers}} \\
\midrule
BERT4Rec \cite{sun2019bert4rec} & 0.415 & 0.479 & 0.706 & 0.182 & 0.578  & 0.502  & 0.560 & 0.796  & 0.128  & 0.231  \\
STAR \cite{lee2025starsimpletrainingfreeapproach} & 0.384 & 0.457 & 0.697 & 0.170 & 0.541 & 0.500 & 0.562 & 0.797 & 0.129 & 0.234 \\
\midrule

\multicolumn{11}{c}{\textbf{Zero-Shot LLMs}} \\
\midrule
GPT-4o      & 0.376 & 0.451 & 0.695 & 0.154 & 0.505  & 0.508 & 0.564 & 0.800  & 0.139  & 0.244  \\
Qwen2.5-3B-Instruct  & 0.375 & 0.451 & 0.693 & 0.160 & 0.515 & 0.497 & 0.553 & 0.794 & 0.131 & 0.245  \\
\midrule

\multicolumn{11}{c}{\textbf{Post-Trained LLMs (Sparse)}} \\
\midrule
Qwen2.5-3B-Instruct & & & & & & & & & & \\
+ TALLRec \cite{Bao_2023}  &  \underline{0.507} & \underline{0.558} & \underline{0.730} & \underline{0.264} & \underline{0.729} &  \underline{0.569} & 
\underline{0.609} & \underline{0.821} & \underline{0.184} & 0.330  \\
+ Rec-R1 \cite{lin2025recr1bridginggenerativelarge}   &   0.391 & 0.466 & 0.700 & 0.174 & 0.545 & 0.554 & 0.600 & 0.817 & 0.170 & 0.325 \\
+ Rank-GRPO \cite{zhu2025rankgrpotrainingllmbasedconversational}   & 0.389 & 0.462 & 0.698  & 0.174  & 0.520  & 0.492 & 0.549 & 0.792 & 0.131 & 0.245  \\
\textbf{+ FlexRec} & \textbf{0.597} & \textbf{0.635} & \textbf{0.788} & \textbf{0.335} & \underline{0.840} & \textbf{0.615} & \textbf{0.665} & \textbf{0.844} & \textbf{0.235} & \textbf{0.368} \\
\textcolor{red}{\small $\Delta$ vs Qwen2.5-3B-Instruct} 
& \textcolor{red}{\small +59.2\%}
& \textcolor{red}{\small +40.8\%}
& \textcolor{red}{\small +13.7\%}
& \textcolor{red}{\small +109.4\%}
& \textcolor{red}{\small +63.1\%}
& \textcolor{red}{\small +23.7\%}
& \textcolor{red}{\small +20.3\%}
& \textcolor{red}{\small +6.3\%}
& \textcolor{red}{\small +79.4\%}
& \textcolor{red}{\small +50.2\%} \\

\bottomrule
\end{tabular}
}
\label{tab:main}
\end{table*}

Table~\ref{tab:main} reports performance on the \textbf{Maximizing Interest} need for KuaiRec and MovieLens-1M. We compare against a broad set of baselines, including traditional rerankers, zero-shot prompted LLMs, and RL post-training methods for LLM rankers. FlexRec consistently outperforms these alternatives by leveraging fine-grained item-level rewards and stabilizing training with uncertainty-aware updates, enabling efficient adaptation of LLM rankers to user needs.

\paragraph{FlexRec Generalizes Across Needs} 


\begin{table*}[h]
\centering
\small
\caption{Need generalization results (train: \textbf{Maximizing Interest}, test: \textbf{Explore New Topics}) indicate that FlexRec transfers strong ranking abilities effectively across needs.
}.
\resizebox{1.0 \linewidth}{!}{%
\begin{tabular}{l|ccccc|ccccc}
\toprule
 & \multicolumn{5}{c|}{\textbf{KuaiRec}} & \multicolumn{5}{c}{\textbf{MovieLens-1M}} \\
\multicolumn{1}{c|}{\textbf{Generalization Setting}}
& \multicolumn{5}{c|}{\textbf{Train: Max-Interest $\rightarrow$ Test: Explore New Topics}}
& \multicolumn{5}{c}{\textbf{Train: Max-Interest $\rightarrow$ Test: Explore New Topics}} \\
\multicolumn{1}{c|}{\textbf{Models}}
& \textbf{N@5} & \textbf{N@10} & \textbf{N@30} & \textbf{R@5} & \textbf{MRR@5}
& \textbf{N@5} & \textbf{N@10} & \textbf{N@30} & \textbf{R@5} & \textbf{MRR@5} \\
\midrule
\multicolumn{11}{c}{\textbf{Traditional Rerankers}} \\
\midrule
BERT4Rec \cite{sun2019bert4rec} & 0.720 & \textbf{0.784} & 0.900 & \textbf{0.165} & \textbf{0.480} & \underline{0.731} & \underline{0.800} & \underline{0.922} &  \underline{0.144} & \underline{0.237}  \\
STAR \cite{lee2025starsimpletrainingfreeapproach} & 0.698 & 0.751 & 0.909 & 0.111 & 0.318 & 0.719 & 0.791 & 0.919 & 0.098 &  0.158\\

\midrule
\multicolumn{11}{c}{\textbf{Zero-Shot LLMs}} \\
\midrule
GPT-4o      &  0.637 & 0.703 & 0.892 & 0.059 & 0.142 & 0.716 & 0.791 & 0.918 & 0.111 & 0.148  \\
Qwen2.5-3B-Instruct  & 0.706 & 0.737 & 0.911 & 0.140 & 0.377 & 0.729  & 0.798  & 0.922  & 0.133 & 0.216 \\
\midrule
\multicolumn{11}{c}{\textbf{Post-Trained LLMs (trained on Max-Interest)}} \\
\midrule
Qwen2.5-3B-Instruct & & & & & & & & & & \\
+ TALLRec \cite{Bao_2023}        &  0.700 & 0.748 & 0.901 & 0.147 & 0.399  & 0.729 & 0.798 & 0.921 & 0.142 & 0.205 \\
+ Rec-R1 \cite{lin2025recr1bridginggenerativelarge}       & \underline{0.730}  & 0.755 & \underline{0.915} & 0.160 & 0.429  & 0.726 & 0.796 & 0.921 & 0.126 & 0.201  \\
+ Rank-GRPO \cite{zhu2025rankgrpotrainingllmbasedconversational}  & 0.717 & 0.767 & \underline{0.915} & \underline{0.163}  & 0.429  & 0.729 & 0.798 & 0.921 & 0.136 & 0.216  \\
\textbf{+ FlexRec}  & \textbf{0.733} & \underline{0.776} & \textbf{0.919} & \textbf{0.165} & \underline{0.440} & \textbf{0.748} & \textbf{0.844} & \textbf{0.944} & \textbf{0.165} & \textbf{0.267} \\

\textcolor{red}{\small $\Delta$ vs Qwen2.5-3B-Instruct}
& \textcolor{red}{\small +3.8\%}
& \textcolor{red}{\small +5.3\%}
& \textcolor{red}{\small +0.9\%}
& \textcolor{red}{\small +17.9\%}
& \textcolor{red}{\small +16.7\%}
& \textcolor{red}{\small +2.6\%}
& \textcolor{red}{\small +5.8\%}
& \textcolor{red}{\small +2.4\%}
& \textcolor{red}{\small +24.1\%}
& \textcolor{red}{\small +23.6\%}
\\
\bottomrule
\end{tabular}
}
\label{tab:gen_explore_topics}
\end{table*}

Table~\ref{tab:gen_explore_topics} and \ref{tab:gen_trend_promotion} show zero-shot generalization performance of baseline models and post-trained LLMs. FlexRec showcases strong need transfer: trained only on \textbf{Maximizing Interest}, our method consistently improves zero-shot performance on \textbf{Explore New Topics} and \textbf{Trend Promotion}. This indicates that fine-grained, need-aware credit assignment helps the model learn reusable ranking principles rather than overfitting to a single objective. Notably, GPT-4o is competitive on MovieLens-1M under \textbf{Trend Promotion}, but FlexRec still achieves the best overall generalization among open-source post-trained counterparts across datasets and settings.

\paragraph{FlexRec is a Universal Multi-Need Ranker}

\begin{figure}
    \centering
    \includegraphics[width=1.0\linewidth]{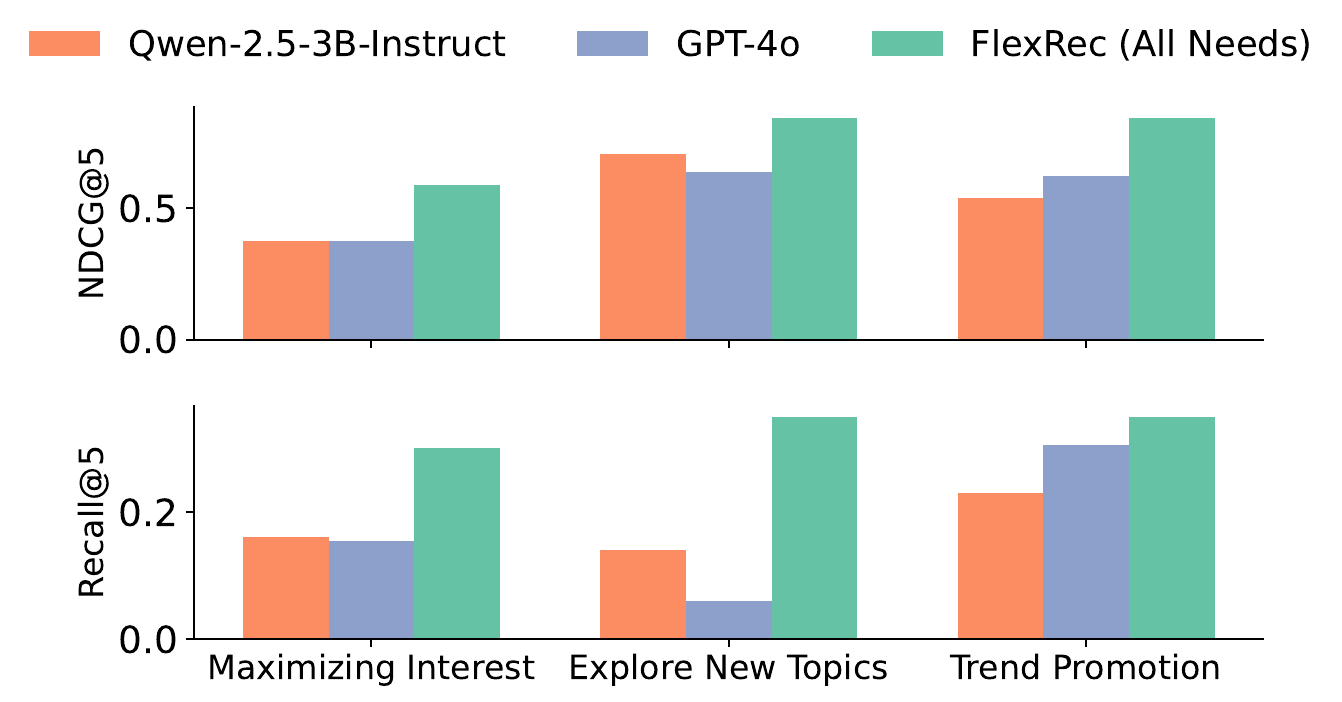}
    \caption{Performance across all three needs on KuaiRec. FlexRec is trained jointly on all needs. Joint training yields consistently stronger performance across needs, supporting FlexRec as an all-purpose recommender conditioned by need instructions.}
    \label{fig:all_needs}
\end{figure}

Figure~\ref{fig:all_needs} and model reasoning in Appendix \ref{appendix:qualitative} shows that FlexRec trained jointly on all needs can internalize heterogeneous objectives and switch ranking behavior at inference via the need instruction. Joint training encourages the model to learn diverse general ranking principles rather than overfitting to a single criterion. Overall, FlexRec represents a practical universal LLM ranker, where one post-trained model serves multiple needs through lightweight prompts.

\subsection{Ablation Studies}

\noindent\textbf{Item-level Reward Formulation}
\label{ablat:item-level}
Table \ref{ablation:reward} shows the performance of different reward constructions under both GRPO and PPO. \emph{Independent contribution} uses an item’s NDCG contribution as reward and is independent of the current candidate pool. \emph{Swap-based contribution (non-causal)} computes rewards via counterfactual swaps, but swaps with both previously predicted (past) and unpredicted (future) items, rather than restricting swaps to the remaining unpredicted pool. Across RL algorithms and reward formulations, our causally grounded swap-based reward yields substantial gains, offering an efficient dense reward for LLM ranking.

\noindent\textbf{Uncertainty-aware Update}
\label{ablat:uncertainty}
Table~\ref{ablat:uncertainty} shows results of methods that address user-interaction sparsity. Traditional recommendation systems use collaborative filtering signals (Item/User KNN) to impute potential interactions. However, they are not optimized to serve as training rewards and provide no estimation of uncertainty, which can yield noisy supervision and brittle updates. In contrast, the learned critic (refer to Appendix \ref{appendix:critic} for details) provides a stronger reward signal, as evidenced by the large gain in performance compared with CF-based methods.Moreover, by explicitly predicting uncertainty and using it to down-weight spurious updates, FlexRec achieves consistent gains in ranking performance across experimental setups.

\subsection{Discussion}

\noindent\textbf{Adaptive Reasoning Across Needs}

As shown in Appendix~\ref{appendix:qualitative}, FlexRec’s fine-grained rewards improve not only ranking quality but also the model’s ability to adapt its reasoning to the active need. For Explore New Topic, the model explicitly identifies the genres/topics present in the user’s history and then seeks complementary, underexposed themes. For Trend Promotion, it performs lightweight numerical reasoning over recent watch counts to prioritize items that are both relevant and currently popular. Such need-conditioned rationales help explain why items are recommended under different needs and represent a step toward more trustworthy and transparent recommendation.

\noindent\paragraph{Impact of Item-level Reward on Training Dynamics}

\begin{figure}[h]
    \centering
    \includegraphics[width=1.0\linewidth]{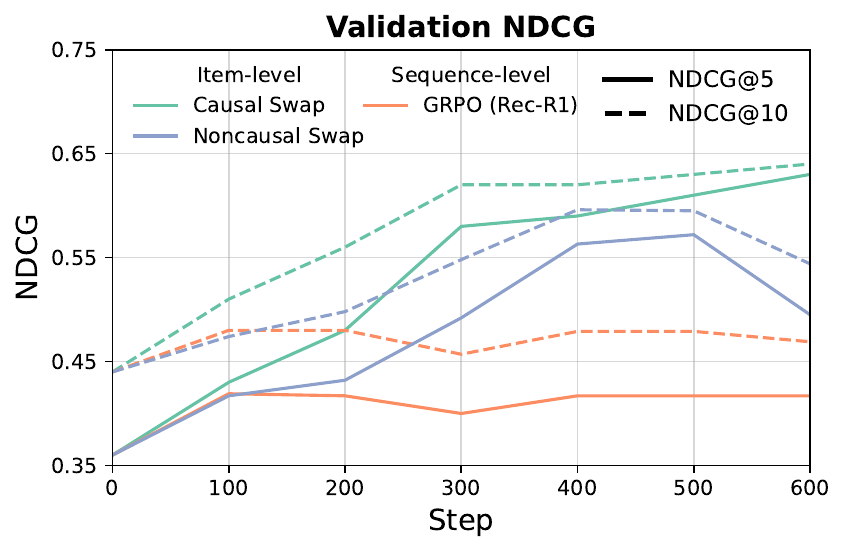}
   \caption{Validation NDCG@5 and NDCG@10 during training under different reward designs, showing that FlexRec's swap-based item-level reward outperforms sequence-level GRPO.}
       \label{fig:discussion_item_level}
\end{figure}

    Figure~\ref{fig:discussion_item_level} compares validation NDCG under different reward designs. The sequence-level reward used in Rec-R1 provides coarse supervision, causing early plateau and weaker performance. In contrast, our swap-based item-level reward delivers fine-grained, position-aware feedback that sustains improvement throughout training. Moreover, allowing swaps with previously selected items in the non-causal variant violates our causality constraint and reduces performance, reinforcing that causally-grounded rewards are critical for autoregressive ranking.

\paragraph{Impact of Uncertainty-aware Updates on Training Dynamics}

\begin{figure}
    \centering
    \includegraphics[width=1.0\linewidth]{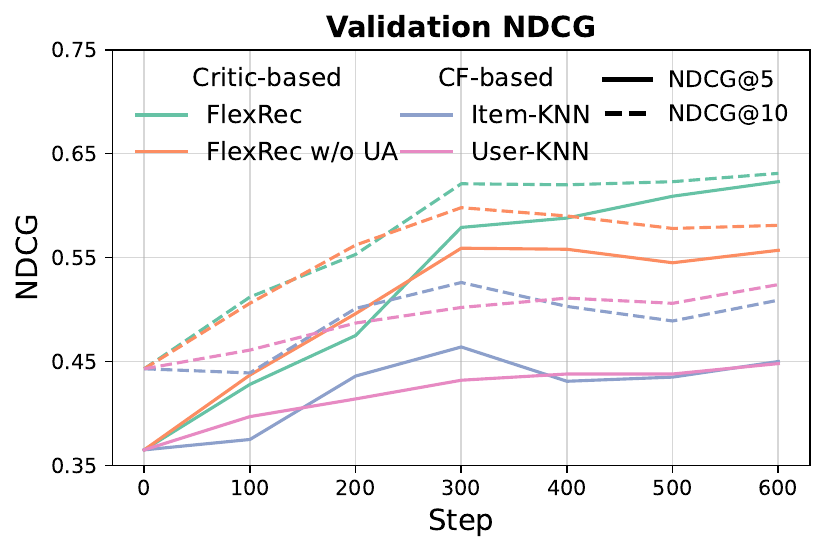}
     \caption{Validation NDCG@5 and NDCG@10 during training under different reward signals (UA denotes uncertainty-aware update), showing that a learned critic with uncertainty-aware GRPO achieves the strongest and most stable learning.}
    \label{fig:discussion_UA}
\end{figure}

Figure~\ref{fig:discussion_UA} shows validation NDCG for models trained with different reward signals. Critic-based FlexRec consistently outperforms KNN-style collaborative-filtering rewards (Appendix~\ref{appendix:critic}), indicating that interaction patterns in sparse feedback are not well captured by CF signals alone. Removing the uncertainty-aware update yields strong early gains but degrades later. We attribute this to the fact that critic error is accumulating and introducing increasingly spurious updates. FlexRec’s uncertainty-aware update mitigates this by downweighting high-variance updates, preserving performance in later training. Overall, the results highlight the importance of modeling and leveraging uncertainty in RLVR under sparse feedback.


\begin{table}[!h]
\centering
\caption{Performance on \textbf{Amazon Product Search} (BERT4Rec not applicable to non-sequential setting), showing that FlexRec is effective not only for sequential recommendation but also for non-sequential ranking tasks such as product search.}

\label{tab:product_search}
\resizebox{\columnwidth}{!}{%
\begin{tabular}{lccccc}
\toprule
\textbf{Product Search}
& \textbf{N@5} & \textbf{N@10} & \textbf{N@30} & \textbf{P@5} & \textbf{R@5} \\
\midrule
\multicolumn{6}{c}{\textbf{Traditional Rerankers}} \\
\midrule
BERT4Rec \cite{sun2019bert4rec} & --- & --- & --- & --- & --- \\
STAR \cite{lee2025starsimpletrainingfreeapproach}    & 0.412 & 0.466  & 0.634  & 0.220 & 0.533  \\
\midrule
\multicolumn{6}{c}{\textbf{Zero-Shot LLMs}} \\
\midrule
GPT-4o      & 0.502 & 0.530 & 0.675 & 0.264 & 0.647  \\
Qwen2.5-3B-Instruct  & 0.449 & 0.493 & 0.650 & 0.235 & 0.585  \\
\midrule
\multicolumn{6}{c}{\textbf{Post-Trained LLMs}} \\
\midrule
Qwen2.5-3B-Instruct & & & & & \\
+ TALLRec \cite{Bao_2023}        & 0.399 & 0.459 & 0.626 & 0.210 & 0.507  \\
+ Rec-R1 \cite{lin2025recr1bridginggenerativelarge}      & 0.504 & 0.536 & 0.679 & 0.268 & 0.652  \\
+ Rank-GRPO \cite{zhu2025rankgrpotrainingllmbasedconversational}  & 0.397 & 0.454 & 0.623 & 0.213 & 0.513  \\
+ FlexRec
                        & \textbf{0.528} & \textbf{0.553} & \textbf{0.690} & \textbf{0.277} & \textbf{0.678}  \\ 
\textcolor{red}{\small $\Delta$ vs Qwen2.5-3B-Instruct}
                        & \textcolor{red}{\small +17.6\%}
                        & \textcolor{red}{\small +12.2\%}
                        & \textcolor{red}{\small +6.2\%}
                        & \textcolor{red}{\small +17.9\%}
                        & \textcolor{red}{\small +15.9\%} \\
\bottomrule
\end{tabular}}
\end{table}



%% file: 05_conclusion.tex
\section{Conclusion}

In this work, we propose FlexRec, an RL post-training framework that adapts large language models to diverse and flexible recommendation needs. By combining a swap-based item-level reward with a variance-aware update, FlexRec mitigates data sparsity and improves the efficiency of credit assignment beyond sequence-level training. Across multiple datasets and needs, FlexRec achieves state-of-the-art performance, highlighting its promise as a universal, general-purpose LLM recommender.

\paragraph{Limitations.}
We study closed-set reranking with predefined candidate pools and labels derived from existing signals, which abstracts away key components of real-world recommenders (e.g., retrieval and open-world item dynamics). Extending this paradigm to more general settings—such as retrieval-augmented personalization and explicitly modeling larger, evolving item spaces—is an important direction for future work.

\section*{Impact Statement}

This paper presents work whose goal is to advance the field of Machine Learning. There are many potential societal consequences of our work, none of which we feel must be specifically highlighted here.

%% file: 10_appendix.tex
\appendix
\onecolumn
\section{Dataset Details}
\label{appendix:data_details}

\paragraph{Data Subsampling}
To ensure scalable training and evaluation, and to facilitate future extensions across datasets and tasks, we subsample each training set to 5,000 instances and each test set to 1,000 instances for all datasets and recommendation settings. 

\paragraph{KuaiRec (short-video recommendation).}
KuaiRec is a fully observed short-video recommendation dataset with dense user--item interactions and rich user/video metadata. Since the raw data is near-dense, we induce sparsity to better match realistic implicit-feedback regimes by uniformly subsampling $10\%$ of interactions per user. We then construct user-level splits with an 80\%/10\%/10\% train/validation/test partition. For each split, we build sequential ranking instances by conditioning on a user’s most recent $H{=}10$ historical items and asking the model to rank a candidate set of $C{=}30$ items; the candidate construction procedure is described in Appendix~\ref{appendix:candidate_generation}. 

\paragraph{MovieLens (rating-based recommendation).}
MovieLens is a classic benchmark for rating-based recommendation that captures long-term user preferences over movies. We use the same dataset splitting and processing protocol as used in KuaiRec.

\paragraph{ESCI (product search).}
ESCI represents product search, where the goal is to retrieve items most relevant to a textual query rather than a user’s interaction sequence. We follow the dataset’s publicly released train--test split for training and evaluation.\footnote{\url{https://github.com/amazon-science/esci-data}}

\section{Need Construction}
\label{appendix:need_construction}
\subsection{Maximizing Interest}

For KuaiRec, following prior work \cite{gao2022kuairec,yang2023generate,10.1145/3589334.3645517}, we use the watch ratio—defined as the ratio between the total watch time and the video length—as a proxy for user interest.
Watch ratio provides a continuous, fine-grained signal that reflects both user engagement and early abandonment behavior, and is commonly adopted as an implicit relevance measure in short-video recommendation settings.
We directly use the watch ratio value as the relevance score when evaluating item utility.

For ML-1M, explicit user feedback is available in the form of discrete ratings ranging from 1 to 5.
We treat higher ratings as stronger expressions of user interest and transform ratings into relevance scores via the exponential mapping $2^{\text{rating}} - 1$.

We use the following instruction across datasets:

\begin{PromptGrayBox}{Need: Maximizing Interest}
 Rank them to maximize the user's expected watch ratio
(likelihood of long/complete viewing) based on their history—because higher watch ratio signals stronger engagement
and improves retention. Also keep in mind to recommend items that strongly align with the user's known interests.
\end{PromptGrayBox}

\subsection{Relevant Novelty}

Niche discovery aims to recommend items that expose users to previously unseen topical items while remaining relevant to their interests.

In KuaiRec, we implement niche discovery using video topic tags.
For a given user history, we identify the set of topics the user has previously interacted with, and label a candidate video as \emph{niche} if its topic does not appear in this history.
Candidates without valid topic annotations are treated as non-niche.

To construct a ground-truth ranking that balances relevance and novelty, we start from a base relevance score obtained via embedding-based retrieval.
We then apply a multiplicative niche bonus to candidates labeled as niche,
\begin{equation*}
\mathrm{score}_{\text{niche}}(v)
=
\mathrm{score}_{\text{base}}(v)\cdot(1+\alpha),
\end{equation*}
where $\alpha>0$ controls the strength of the novelty preference.
The resulting scores induce a ranking that prioritizes novel-topic items while preserving relevance among both niche and non-niche candidates.
This formulation explicitly counteracts exposure bias by amplifying under-explored items without discarding similarity-based relevance.

For ML-1M, niche discovery is defined analogously by identifying movies whose genres have not appeared in the user’s interaction history.

We use the following instruction across datasets: 

\begin{PromptBlueBox}{Need: Niche Discovery}
Keep in mind to recommend not only items aligned with the user’s
interests but also items that are niche (topics the user hasn't interacted with), but that are plausibly relevant
and could enrich their experience.
\end{PromptBlueBox}

\subsection{Trend Promotion}

Trending recommendation aims to surface items that are both relevant to the user and currently popular across the platform, capturing short-term collective interest and temporal dynamics.

In KuaiRec, we construct trending signals using recent interaction frequency.
For each candidate video, we compute its trend score as the number of user interactions occurring within a sliding \textbf{one-day} window prior to the query time.

To construct a ground-truth ranking that balances personalization and global popularity, we combine semantic relevance and temporal trend signals.
Candidate videos are first retrieved using embedding-based similarity to the user’s recent interaction history.
Both the similarity scores and the one-day trend counts are independently normalized via min–max normalization.
The final relevance score is computed as the combination,
\begin{equation}
\mathrm{score}_{\text{trend}}(v)
=
\alpha \cdot \mathrm{sim}_{\text{norm}}(v)
+
(1-\alpha) \cdot \mathrm{trend}_{\text{norm}}(v),
\end{equation}
where \(\alpha \in (0,1)\) controls the trade-off between personalized relevance and global trending popularity (we use $\alpha = 0.7$ across experiments).
The resulting scores induce a ranking that prioritizes items that are both timely and aligned with the user’s interests.

For ML-1M, trending recommendation is defined analogously using recent 1-day activity.

We use the following instruction across datasets: 

\begin{PromptBrownBox}{Need: Trending Promotion}
Keep in mind to recommend not only items aligned with the user’s interests but also items that are popular in the last 24 hours (provided as the number of watches in the last day), as long as they are relevant to the user.
\end{PromptBrownBox}

\subsection{Product Search}

Amazon’s official ESCI dataset annotates each (query, item) pair into four relevance categories—\textbf{Exact (E)}, satisfying all query constraints; \textbf{Substitute (S)}, partially matching but usable as a replacement; \textbf{Complement (C)}, not matching the query itself but useful alongside an exact item; and \textbf{Irrelevant (I)}. We follow these categories and the official graded scoring scheme, mapping them to \(\{1.0,\,0.1,\,0.01,\,0.0\}\) for \(\{E,S,C,I\}\), respectively.

We use the following instruction for product search: 

\begin{PromptTealBox}{Need: Product Search}
Your task is to rerank these candidate products by how well they match the user's search query. Consider how closely each product's title, description, and attributes align with the query intent. Give higher ranks to products that are exact or highly relevant matches (e.g., correct type, brand, or specifications). Lower the rank of products that are complementary or unrelated to the query.
\end{PromptTealBox}

\subsection{Data Subsampling}
To ensure scalable training and evaluation, and to facilitate future extensions across datasets and tasks, we subsample each training set to 5,000 instances and each test set to 1,000 instances for all datasets and recommendation settings. 

\section{RL Training Details}
\label{app:rl_training_details}

We perform reinforcement learning (RL) post-training on {4$\times$ NVIDIA A100 GPUs using the Verl \footnote{\url{https://github.com/verl-project/verl}} library. Unless otherwise specified, all experiments use full-parameter training with learning rate $1\times 10^{-6}$.

To stabilize optimization and constrain policy drift, we include a KL regularization term against a reference policy with KL loss coefficient $0.01$. We additionally apply entropy regularization to encourage exploration and avoid premature collapse, using entropy coefficient $0.005$.

For rollout generation, we use \textbf{vLLM} as the inference backend to efficiently sample sequences during RL training, improving throughput and reducing end-to-end training latency.

\section{Candidate Generation}
\label{appendix:candidate_generation}

We construct the candidate set using a fixed semantic retriever following STAR~\cite{lee2025starsimpletrainingfreeapproach}, a state-of-the-art training-free retrieval framework based on large language model embeddings.
Each item is first converted into a natural-language description using its available metadata, and embedded using the \texttt{text-embedding-004} model from the Gemini embedding family \cite{lee2024geckoversatiletextembeddings}.

Given a user interaction history of length $H=10$, candidate items are retrieved by cosine similarity in the embedding space.
To account for temporal preference dynamics, we form a query embedding as a temporally discounted average of historical item embeddings, optionally weighted by interaction strength (e.g., ratings or watch ratios), following the original STAR formulation.
The candidate set consists of the top-$K$ most similar items, excluding those already interacted with by the user.

The retriever is kept frozen throughout training and evaluation.
This setup reflects realistic recommendation pipelines, where candidate generation is typically performed by a pre-trained, independently maintained retrieval system, while downstream ranking models are optimized on a fixed candidate pool.

\section{Item Representation}

In this section we detail the relevant information we give to the models. We give the same set of metadata for embedding models in retrieval as well as for LLMs in reranking.

\subparagraph{Videos (KuaiRec).}
For short videos, we leverage rich categorical and item annotations provided by the dataset.
Each video is converted into a textual description composed of its caption, hierarchical category path, and optional tag information:
\begin{quote}
\small
\texttt{Video caption: <caption>} \\
\texttt{Category path: <level\_1> > <level\_2> > <level\_3>} \\
\texttt{Video tag: <tag>}
\end{quote}

\subparagraph{Movies (MovieLens-1M).}
For movies, we construct a concise textual description consisting of the movie title, release year, and genre information:
\begin{quote}
\small
\texttt{Movie Title: <title>} \\
\texttt{Movie Year: <year>} \\
\texttt{Movie genres: <genre\_1>, <genre\_2>, \dots}
\end{quote}

\subparagraph{ESCI}
For the ESCI product search dataset, each product is represented using core catalog metadata commonly available in real-world e-commerce systems.
Specifically, we convert each item into a textual description by concatenating its product title, bullet-point description, brand, and color attributes:
\begin{quote}
\small
\texttt{Product title: <product\_title>} \\
\texttt{Product description: <product\_bullet\_point>} \\
\texttt{Brand: <product\_brand>} \\
\texttt{Color: <product\_color>}
\end{quote}
This representation captures both functional semantics (via title and description) and distinguishing attributes (via brand and color), which are critical signals for product relevance in search scenarios.
All fields are expressed in natural language and fused into a single text sequence prior to embedding, enabling a unified semantic retrieval interface consistent with other datasets.

\section{Evaluation Details}
\label{appendix:evaluation_details}
\paragraph{Addressing Invalid Output}
Prior work~\cite{ren2025learningdynamicsllmfinetuning,deng2025onerec} shows that negative advantage can induce invalid outputs by ``squeezing'' the policy out of valid regimes. We observe a similar trend: as models improve ranking quality, they may also emit invalid item IDs. We view this as expected behavior: the model should avoid low-quality items. Thus, we treat invalid IDs as low-confidence predictions. For fair evaluation, when an output contains fewer than $K$ valid items, we randomly pad the list with the rest candidate items to reach length $K$. This padding has negligible impact on our metrics, since they are dominated by the top-ranked items.

\section{Interaction Critic and Uncertainty Estimation}
\label{appendix:critic}
\begin{table*}[h]
\centering
\small
\caption{Critic performance on engagement prediction (watch ratio for KuaiRec and rating for ML-1M). We report mean prediction accuracy and uncertainty quality measured by the correlation between predicted variance and squared prediction error. An asterisk indicates statistical significance ($p<0.05$).}
\setlength{\tabcolsep}{6pt}
\begin{tabular}{l|cccc|cccc}
\toprule
& \multicolumn{4}{c|}{\textbf{KuaiRec}} 
& \multicolumn{4}{c}{\textbf{ML-1M}} \\
\textbf{Method}
& \textbf{MSE} $\downarrow$
& \textbf{MAE} $\downarrow$
& \textbf{Pearson (Mean)} $\uparrow$
& \textbf{Pearson (Var)} $\uparrow$
& \textbf{MSE} $\downarrow$
& \textbf{MAE} $\downarrow$
& \textbf{Pearson (Mean)} $\uparrow$
& \textbf{Pearson (Var)} $\uparrow$ \\
\midrule

\textbf{Critic}
& 0.328 & 0.400 & $0.374^{*}$ & $0.300^{*}$ 
& 0.317 & 0.271 & $0.510^{*}$ & $0.132^{*}$  \\
\bottomrule
\label{appendix:critic_result}
\end{tabular}
\label{tab:critic_results}
\end{table*}
\paragraph{Task Definition and Challenges}
Given a user’s interaction context which consists of historical items and observed engagement signals, we aim to predict the user’s potential interaction with a candidate item. This prediction task is inherently challenging, subsuming the reranking problem considered in this work. Accordingly, the goal of the critic is not to produce perfectly accurate engagement predictions, but rather to extract and provide useful learning signals that can be exploited during the reinforcement learning stage.

In recommendation settings, such engagement signals (e.g., watch ratio or dwell time) exhibit two properties.
First, they are \emph{long-tailed}, with a small fraction of interactions accounting for a disproportionate share of total engagement.
Second, they are \emph{heteroscedastic}: the intrinsic noise of user feedback varies substantially across contexts, users, and items.
As a result, both the expected engagement level and the uncertainty of this estimate depend on the input context.

\paragraph{Critic Architecture}
We employ a two-tower MLP-based critic (see \ref{appendix:critic_architecture} for details) that encodes user history and candidate items into a shared embedding space, followed by a lightweight interaction network.
The critic outputs both a predicted mean engagement $\hat{\mu}(X)$ and an input-dependent variance $\sigma^2(X)$.
The architecture is intentionally simple, as the critic serves as a training signal rather than a primary ranking model.

\begin{table}[t]
\centering
\small
\caption{Architecture of the critic model used to estimate engagement mean and uncertainty. where $\hat{u}$ represents (learnt) user embedding and $\hat{c}$ represents (fixed) candidate item embedding.}
\begin{tabular}{l c}
\toprule
\textbf{Component} & \textbf{Specification} \\
\midrule
User history encoder & MLP with ReLU activations \\
Candidate encoder & Fixed item embeddings \\
Hidden dimension & 256 \\
Interaction features & $\hat{u}, \hat{c}, \hat{u}\odot\hat{c}, \hat{u}^\top\hat{c}$ \\
Output heads & Mean head + log-variance head \\
Normalization & LayerNorm \\
Dropout & 0.05 \\
\bottomrule
\end{tabular}
\label{appendix:critic_architecture}
\end{table}

\paragraph{Variance-Weighted Gaussian Likelihood}
To model heteroscedastic engagement noise, we use recent advances in heteroscedastic regression using neural networks \cite{seitzer2022pitfallsheteroscedasticuncertaintyestimation}.
Let $Y$ denote the observed engagement signal and $X$ the input context.
The critic predicts a mean $\hat{\mu}(X)$ and variance $\sigma^2(X)$.
We minimize the following loss:
\begin{equation}
\mathcal{L}_{\beta\text{-NLL}}
=
\mathbb{E}_{(X,Y)}\!\left[
\sigma^2(X)\left(
\frac{1}{2}\log \sigma^2(X)
+
\frac{(Y-\hat{\mu}(X))^2}{2\sigma^2(X)}
\right)
\right].
\end{equation}

We follow the original work and use $\beta = 1$

\paragraph{Training}

On both KuaiRec and ML-1M, we first conduct a 8:1:1 split on all interactions data. Then we train the critic for 30 epochs and pick the checkpoint with the best MSE.

\paragraph{Results}

Table~\ref{appendix:critic_result} reports the critic’s performance. While the predicted values are positively correlated with the ground-truth scores—suggesting the critic can provide useful reward signals for RLVR—they still exhibit substantial variance and nontrivial prediction error. This motivates explicitly modeling uncertainty during RL to downweight unreliable signals and avoid destabilizing updates.

\subsection{Qualitative Analysis on Multiple Needs}
\label{appendix:qualitative}

In this section, we provide a qualitative analysis of how the trained LLM adjusts its reasoning and predictions across a diverse set of needs. As illustrated by the responses under different need instructions, the model dynamically adapts its decision process and produces need-relevant rationales that guide its ranking choices. These examples further suggest that \textsc{FlexRec} can serve as a universal, general-purpose recommender capable of handling heterogeneous objectives within a single model. Finally, we note that the generated reasoning is valuable not only for improving prediction, but also for offering explainability and transparency—properties that are often missing in traditional recommender systems \cite{ge2024surveytrustworthyrecommendersystems}.

\begin{PromptGrayBox}{Need: Maximizing Interest}
\textbf{Prompt:}

Given user history (in chronological order): \ldots\ Rank them to maximize the user's expected watch ratio
(likelihood of long/complete viewing) based on their history—because higher watch ratio signals stronger engagement
and improves retention. Also keep in mind to recommend items that strongly align with the user's known interests.

\medskip
\textbf{Model Response:}

\texttt{\textless think\textgreater}\\
Based on the user's interest in celebrity gossip coupled with a liking for short engaging moments related to popular
celebrities, a strong recommendation would include several videos under `celebrity amusement' category path. The
variety might increase the chances for longer engagement. Another notable behavior could involve following content
lengths of approximately equal duration to what we see the user engaging normally with, at around 11.22 seconds play
duration.\\
\texttt{\textless /think\textgreater}

\texttt{\textless answer\textgreater}: \ldots
\end{PromptGrayBox}

\begin{PromptBlueBox}{Need: Niche Discovery}
\textbf{Prompt:}

Given user history (in chronological order): \ldots\ Keep in mind to recommend not only items aligned with the user’s
interests but also items that are niche (topics the user hasn't interacted with), but that are plausibly relevant
and could enrich their experience.

\medskip
\textbf{Model Response:}

\texttt{\textless think\textgreater}\\
First, let's look at how videos are categorized and tagged. Users seem interested in aesthetics (specific tags like
``beauty''), emotions, entertainment (tags like ``amusement'' and ``short play''), practical life hacks (``life''),
and family events (like weddings and birth). However, there were no videos regarding kids or parenting recently noted,
which could be niche but relevant.\\
\texttt{\textless /think\textgreater}

\texttt{\textless answer\textgreater}: \ldots
\end{PromptBlueBox}

\begin{PromptBrownBox}{Need: Trend Promotion} 
\textbf{Prompt:}

Given user history (in chronological order): \ldots\ Keep in mind to recommend not only items aligned with the user’s interests but also items that are popular in the last 24 hours (provided as the number of watches in the last day), as long as they are relevant to the user. \\

\textbf{Model Response:} \\
\texttt{\textless think\textgreater}\\
I will start with the most watched video in the last 24 hours ('1429'), which is '6549 views'. This should grab attention as it's very popular right now. Next I'll pick videos with high watches, like '4803', '1642' and others. Then I'll include some trending content with a mix of topics and styles. After that, I'll add newer, higher-watched videos along with those in the browsing history, maintaining order from highest to lowest. Finally, I'll round out the list with lower preference but still relevant videos. \\
\texttt{\textless /think\textgreater}

\texttt{\textless answer\textgreater}: \ldots
\end{PromptBrownBox}

\begin{PromptTealBox}{Need: Product Search} 
\textbf{Prompt:}

Given user query: qscreen fence without holes \ldots\ Your task is to rerank these candidate products by how well they match the user's search query. Consider how closely each product's title, description, and attributes align with the query intent. Give higher ranks to products that are exact or highly relevant matches (e.g., correct type, brand, or specifications). Lower the rank of products that are complementary or unrelated to the query. \\

\textbf{Model Response:} \\
\texttt{\textless think\textgreater}\\
I start by examining the user query '!qscreen fence without holes' which seems to indicate a need for a fence that screens off an area but does not require installing holes or special hardware (like grommets or tie downs). Based on this, I'll consider which ASINs would likely match this request- B07DS1YCRZ from ColourTree appears to be a suitable product, offering commercial grade, UV-stabilized fence screening fabric with cable zip ties included. The 'screen' description aligns perfectly with the search criteria.- B08NG85RHL from Orgrimmar is also a heavy-duty privacy screen intended for residential and commercial uses, matching the commercial-grade requirement and screen usage, though its specific size is too large compared to what might be needed for normal fences.
\texttt{\textless /think\textgreater}

\texttt{\textless answer\textgreater}: \ldots
\end{PromptTealBox}

\section{Generalization Results on Trend Promotion}
\begin{table*}[h]
\centering
\small
\caption{Need generalization results (train: \textbf{Maximizing Interest}, test: \textbf{Trend Promotion}) indicate that FlexRec transfers strong ranking abilities effectively across needs.}
\resizebox{0.8 \linewidth}{!}{%
\begin{tabular}{l|ccccc|ccccc}
\toprule
 & \multicolumn{5}{c|}{\textbf{KuaiRec}} & \multicolumn{5}{c}{\textbf{MovieLens-1M}} \\
\multicolumn{1}{c|}{\textbf{Generalization Setting}}
& \multicolumn{5}{c|}{\textbf{Train: Max-Interest $\rightarrow$ Test: Trend Promotion}}
& \multicolumn{5}{c}{\textbf{Train: Max-Interest $\rightarrow$ Test: Trend Promotion}} \\
\multicolumn{1}{c|}{\textbf{Models}}
& \textbf{N@5} & \textbf{N@10} & \textbf{N@30} & \textbf{R@5} & \textbf{MRR@5}
& \textbf{N@5} & \textbf{N@10} & \textbf{N@30} & \textbf{R@5} & \textbf{MRR@5} \\
\midrule

\multicolumn{11}{c}{\textbf{Traditional Rerankers}} \\
\midrule
BERT4Rec \cite{sun2019bert4rec} & 0.479 & 0.520 & 0.774 & 0.123 & 0.085 & 0.542 & 0.601 & 0.820 & 0.181 & 0.212 \\
STAR \cite{lee2025starsimpletrainingfreeapproach} & 0.579 & 0.624 & 0.821 & 0.164 & 0.095 & 0.604 & 0.655 & 0.844 & 0.228 & 0.280 \\
\midrule
\multicolumn{11}{c}{\textbf{Zero-Shot LLMs}} \\
\midrule
GPT-4o      & 0.622 & 0.647 & 0.844 & 0.306 & 0.112 & 0.669 & 0.710 & 0.868 & 0.320  & 0.349 \\
Qwen2.5-3B-Instruct  & 0.540 & 0.582 & 0.808  & 0.230 & 0.176  & 0.581 & 0.627  & 0.834 & 0.220 & 0.280 \\
\midrule
\multicolumn{11}{c}{\textbf{Post-Trained LLMs (trained on Max-Interest)}} \\
\midrule
Qwen2.5-3B-Instruct & & & & & & & & & & \\

+ TALLRec \cite{Bao_2023}        & 0.493 & 0.552 & 0.790 & 0.152 & 0.118  & 0.611 & 0.648 & 0.845 & 0.244 & 0.322  \\
+ Rec-R1  \cite{lin2025recr1bridginggenerativelarge}     & 0.498 & 0.552 & 0.791 & 0.158 & 0.120 & 0.607 & 0.653 & 0.844 & 0.252  & 0.306  \\
+ Rank-GRPO \cite{zhu2025rankgrpotrainingllmbasedconversational}  & 0.472 & 0.531 & 0.780 & 0.115 & 0.080  & 0.510 & 0.573 & 0.808 & 0.139 & 0.184  \\
\textbf{+ FlexRec}
                        & 0.563 & 0.611 & 0.816 & 0.167 & 0.124 
                        & 0.630 & 0.677 & 0.854  & 0.269  & 0.329  \\
\textcolor{red}{\small \% Gain vs Qwen2.5-3B-Instruct}
& \textcolor{red}{\small +4.3\%}
& \textcolor{red}{\small +5.0\%}
& \textcolor{red}{\small +1.0\%}
& \textcolor{gray!60}{\small -27.4\%}
& \textcolor{gray!60}{\small -29.5\%}
& \textcolor{red}{\small +8.4\%}
& \textcolor{red}{\small +8.0\%}
& \textcolor{red}{\small +2.4\%}
& \textcolor{red}{\small +22.3\%}
& \textcolor{red}{\small +17.5\%}
\\

\bottomrule
\end{tabular}
}
\label{tab:gen_trend_promotion}
\end{table*}

\begin{table*}[t]
\centering
\small
\caption{Ablation of reward formulations for GRPO and PPO on the KuaiRec \emph{Maximizing Interest} task under the dense-interaction setting.}
\label{ablation:reward}

\setlength{\tabcolsep}{4pt}
\begin{tabular}{l l c c c c c}
\toprule
\textbf{Alg.} & \textbf{Reward Formulation}
& \textbf{N@5} & \textbf{N@10} & \textbf{N@30} & \textbf{R@5} & \textbf{MRR@5} \\
\midrule

\multirow{3}{*}{GRPO}
& Independent contribution                    & 0.380 & 0.455 & 0.695 & 0.167 & 0.533  \\
& Swap-based contribution (non-causal)        & 0.461 & 0.514 & 0.731 & 0.214 & 0.716  \\
& \textbf{CausalSwap reward (ours)}            & \textbf{0.607} & \textbf{0.636} & \textbf{0.790} & \textbf{0.336} & \textbf{0.851} \\
\midrule

\multirow{3}{*}{PPO}
& Independent contribution                    & 0.390 & 0.462 & 0.698 & 0.175 & 0.546  \\
& Swap-based contribution (non-causal)        & 0.383 & 0.455 & 0.695  & 0.169  & 0.534 \\
& \textbf{CausalSwap reward (ours)}           & \textbf{0.621} & \textbf{0.661} & \textbf{0.803} & \textbf{0.338}  & \textbf{0.850}  \\
\bottomrule
\end{tabular}
\end{table*}

\begin{table}[t]
\centering
\footnotesize
\caption{Ablation of reward signals. Critic-based rewards outperform CF heuristics, and variance-aware updates further improve performance.}
\setlength{\tabcolsep}{4pt}
\begin{tabular}{l c c c c c}
\toprule
\textbf{Reward Signal}
& \textbf{N@5}
& \textbf{N@10}
& \textbf{N@30}
& \textbf{R@5}
& \textbf{MRR@5} \\
\midrule

\multicolumn{6}{l}{\textit{CF-based}} \\
User-KNN CF  & 0.410  & 0.484 & 0.707 & 0.194 & 0.570  \\
Item-KNN CF  & 0.417  & 0.480 & 0.707  & 0.208  & 0.603  \\

\midrule
\multicolumn{6}{l}{\textit{Critic-based}} \\
Raw critic     & 0.566  & 0.591 &  0.775  & 0.280 & \textbf{0.831}  \\
\textbf{FlexRec: Uncertainty-aware Critic}
               & \textbf{0.595} & \textbf{0.623} & \textbf{0.786} & \textbf{0.319} & 0.825 \\
\bottomrule
\end{tabular}
\label{ablat:uncertainty}
\end{table}

\section{Results on Dense Setting of KuaiRec}
\label{appendix:kuairec_dense}
Throughout this work, we primarily assume sparse user--item interactions, as is typical in practice. KuaiRec has a dense version, where nearly every user has interacted with nearly every item. In this setting, rewards for user--item pairs are largely observed, reducing the need for a learned critic. Table~\ref{appendix:tab_dense_KuaiRec} reports results on KuaiRec-dense. FlexRec still achieves state-of-the-art performance, demonstrating its robustness across both sparse and dense settings.

\begin{table*}[h]
\small
\centering
\caption{Performance on the \textbf{Maximizing Interest} need on dense version of KuaiRec.}
\label{appendix:tab_dense_KuaiRec}
\begin{tabular}{l|ccccc}
\toprule
\textbf{Maximizing Interest (KuaiRec)}
& \textbf{N@5} & \textbf{N@10} & \textbf{N@30} & \textbf{R@5} & \textbf{MRR@5} \\
\midrule

\multicolumn{6}{c}{\textbf{Baselines}} \\
\midrule
BERT4Rec & 0.415 & 0.479 & 0.706 & 0.182 & 0.578 \\
STAR (retriever) & 0.384 & 0.457 & 0.697 & 0.170 & 0.541 \\
GPT-4o      & 0.376 & 0.451 & 0.695 & 0.154 & 0.505 \\
Qwen2.5-3B  & 0.375 & 0.451 & 0.693 & 0.160 & 0.515 \\
\midrule

\multicolumn{6}{c}{\textbf{Post-Trained LLMs (Dense)}} \\
\midrule
Qwen2.5-3B + SFT        & 0.585  & 0.612  & 0.756  & 0.333  & 0.783 \\
Qwen2.5-3B + PPO        & 0.386  & 0.458  & 0.697  & 0.170  & 0.541 \\
Qwen2.5-3B + Rec-R1     & 0.572  & 0.616  & 0.769  & 0.307  & 0.780 \\
Qwen2.5-3B + Rank-GRPO  & 0.381  & 0.456  & 0.696  & 0.168  & 0.517 \\
\textbf{Qwen2.5-3B + GRPO (ours)}
                       & \textbf{0.607} & \textbf{0.636} & \textbf{0.790} & \textbf{0.336} & \textbf{0.851} \\
\bottomrule
\end{tabular}
\end{table*}